\documentclass{article}

\usepackage{PRIMEarxiv}

\usepackage[utf8]{inputenc} %
\usepackage{algorithm}
\usepackage{array}
\usepackage{algpseudocode}
\usepackage{subfig}
\usepackage{textcomp}
\usepackage{stfloats}
\usepackage{verbatim}
\usepackage[T1]{fontenc}    %
\usepackage{hyperref}       %
\usepackage{url}            %
\usepackage{booktabs}       %
\usepackage{amsfonts}       %
\usepackage{nicefrac}       %
\usepackage{microtype}      %
\usepackage{xcolor}         %

\usepackage{amsmath}
\usepackage{bm}
\usepackage{color}
\usepackage{graphicx}
\usepackage{makecell}
\usepackage{multirow}
\usepackage{tabularx}
\usepackage{mathtools}
\usepackage{pythonhighlight}
\usepackage{listings}

\definecolor{pink}{rgb}{0.858, 0.188, 0.478}
\usepackage{xspace}

\definecolor{commentcolor}{RGB}{110,154,155}   %

\title{Contrastive ECOC: Learning Output Codes for Adversarial Defense
}

\author{
  Che-Yu Chou, Hung-Hsuan Chen \\
  Computer Science and Information Engineering \\
  National Central University \\
  Taoyuan, Taiwan\\
  \texttt{tetsuyu89617@gmail.com, hhchen1105@acm.org} \\
}

\begin{document}
\maketitle

\begin{abstract}

Although one-hot encoding is commonly used for multiclass classification, it is not always the most effective encoding mechanism. Error Correcting Output Codes (ECOC) address multiclass classification by mapping each class to a unique codeword used as a label. Traditional ECOC methods rely on manually designed or randomly generated codebooks, which are labor-intensive and may yield suboptimal, dataset-agnostic results. This paper introduces three models for automated codebook learning based on contrastive learning, allowing codebooks to be learned directly and adaptively from data. Across four datasets, our proposed models demonstrate superior robustness to adversarial attacks compared to two baselines. The source is available at \url{https://github.com/YuChou20/Automated-Codebook-Learning-with-Error-Correcting-Output-Code-Technique}.

\end{abstract}


\section{Introduction} \label{sec:intro}

Deep learning has achieved remarkable success in various fields. However, it is susceptible to adversarial examples: subtle perturbations of input data can significantly mislead model predictions~\cite{szegedy2013intriguing}. This vulnerability motivates extensive research into adversarial attack methods~\cite{goodfellow2014explaining,madry2017towards} and defense strategies~\cite{szegedy2013intriguing,goodfellow2014explaining,papernot2016distillation}.

For multiclass classification, conventional one-hot encoding is not always optimal, as it treats classes as orthogonal, thus ignoring inter-class relationships and offering limited flexibility for handling adversarial settings. 

Recent studies have explored Error Correcting Output Codes (ECOCs)~\cite{verma2019error,song2021error,wan2022efficient,philippon2023improved,dietterich1994solving} to enhance the robustness of the model against adversarial attacks like FGSM, PGD, and C\&W~\cite{goodfellow2014explaining,madry2017towards}. ECOCs assign a unique codeword to each class, leveraging error correction principles for improved resilience. However, manually designing codebooks is labor-intensive and often suboptimal for diverse datasets.

To address these limitations, this paper introduces novel methods for automated codebook learning (ACL). We leverage the row separation loss and column separation loss, which are key codebook design principles, to maximize the separation and independence of codewords, enabling the model to generate dataset-specific codebooks that enhance accuracy and robustness against adversarial attacks. Experimental results demonstrate that the proposed models outperform traditional manually designed and randomly generated codebooks. Additionally, the models exhibit superior resilience to adversarial attacks, providing a robust solution for multiclass classification.

\section{Method} \label{sec:method}

\subsection{Codebook Design Philosophy}

\begin{table}[tb]
    \caption{An ECOC codebook example with 4 classes; each codeword length is 10. The last row is predicted as class 4 since it is closest to the 4th class by Hamming distance.}
    \label{tab:codebook-example}
    \centering
    \begin{tabular}{@{}c|cccccccccc@{}}
    \toprule
     & $c_1$ & $c_2$ & $c_3$ & $c_4$ & $c_5$ & $c_6$ & $c_7$ & $c_8$ & $c_9$ & $c_{10}$ \\ \midrule
    $1$                      & 0  & 0  & 0  & 0  & 0  & 0  & 0  & 0  & 0  & 0   \\
    $2$                     & 0  & 1  & 1  & 0  & 1  & 0  & 1  &1  & 1  & 1   \\
    $3$                    & 1  & 1  &0 &1  & 1  & 1  & 0  & 0  & 1  & 1   \\
    $4$                   & 1  & 0  & 1  & 1  & 0  & 1  & 1  & 0  & 0  & 1   \\
    \midrule
    $?$                  & 1  & 0  & 1  & 1  & 0  & 1  & 0  & 1  & 1  & 1   \\ 
    \end{tabular}
\end{table}

Table~\ref{tab:codebook-example} illustrates an example of an ECOC codebook: the top 4 rows show 4 classes; each class has a 10-bit codeword. Each bit is the output of a binary classifier. During prediction, concatenated classifier outputs form a predicted codeword. The class with the minimum Hamming distance to the predicted codeword is chosen. For instance, the last row's predicted codeword is closest to class 4, yielding a prediction of class 4.

Codebook design follows two principles:
First, Row Separation: maximize inter-class codeword distances, which offers greater error tolerance, making predictions robust even with minor errors.
Second, Column Separation: minimize inter-column correlation to prevent simultaneous errors from different classifiers.

We propose three methods for ACL adhering to these principles.

\subsection{ACL-PF: ACL by Pretraining \& Finetuning} \label{sec:model_arch_design}

Our first model, ACL-PF, leverages a two-stage pretraining and fine-tuning strategy, integrating SimCLR~\cite{chen2020simple} with ECOC principles. Table~\ref{tab:symbol-intro} lists the symbols used.

\begin{figure*}[tb]
\centering
\includegraphics[width=\textwidth]{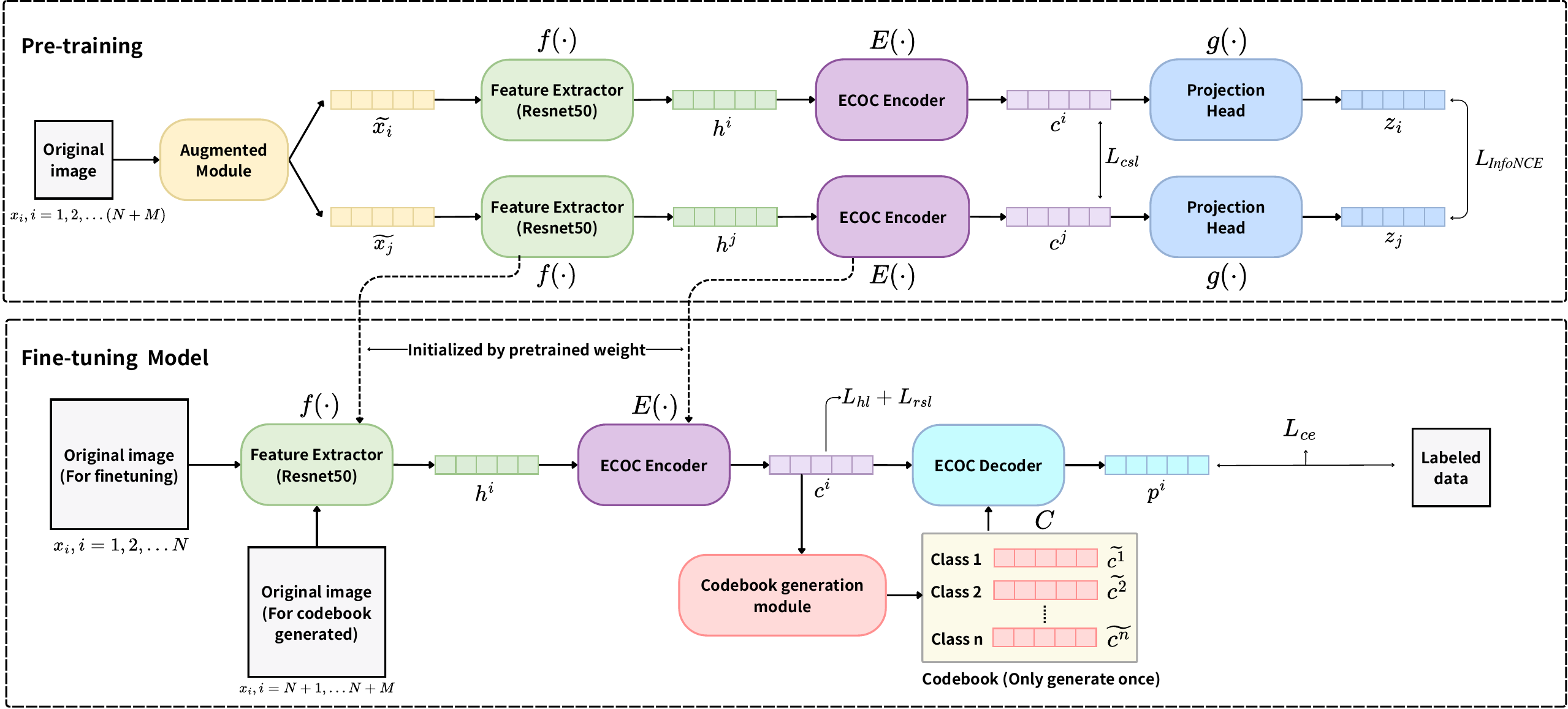}
\caption[ACL-PF Network Architecture]{ACL-PF network architecture. Top: pretraining phase, Bottom: fine-tuning phase.}
\label{fig:model1_arch}
\end{figure*}

\begin{table}[tb]
\caption[Symbols Used in This Paper]{Symbols used in this paper.}
\label{tab:symbol-intro}
\centering
\begin{tabularx}{\columnwidth}{cX}
\toprule
Symbol & Description \\ \midrule
$x_i, y_i$ & Input instance $i$'s features and label, $i=1,\ldots, N$ \\
$\tilde{x}_i$ & $x_i$'s augmentation \\
$h^i$ & $x_i$'s feature representation extracted by $f(\cdot)$ \\
$c^i$ & $x_i$'s predicted codeword \\
$z_i$ & $x_i$'s output vector of projection head $g(\cdot)$ \\
$\tilde{c}^k$ & Codeword for class $k$ in the codebook \\
$\tilde{c}^k_j$ & $j$th element of codeword $\tilde{c}^k$, $j=1,\ldots, n$ \\
$p_k^i$ & Probability of input $x_i$ being class $k$, $k=1,\ldots, K$ \\
$c_{f_j}$ & Output vector of the $j$th base classifier for a batch \\
$M$ & Number of data points in $D_{cg}$ for codebook generation \\
$C$ & Codebook \\
\bottomrule
\end{tabularx}
\end{table}

\subsubsection{Pretraining Stage}
As shown in Figure~\ref{fig:model1_arch}, pretraining mimics contrastive learning (CL) for data representations. Unlike classic CL, we add an ECOC encoder and a column separation loss ($L_{csl}$) to minimize column correlation (Eq.~\ref{eq:csl_def}). Codewords represent classes.

The pretraining uses data augmentation, a feature extractor (ResNet), an ECOC encoder $E(\cdot)$, and a projection head. Data augmentation generates positive pairs $\tilde{x}_i, \tilde{x}_j$. The feature extractor generates input for the ECOC encoder $E(\cdot)$ (a fully connected layer), which maps classes to unique codewords $c^i$, enforcing column separation via $L_{csl}$ (Eq.~\ref{eq:csl_def}). The projection head maps $c^i$ to a space to minimize InfoNCE loss (Eq.~\ref{eq:infonce})~\cite{chen2020simple}.

\begin{equation}\label{eq:csl_def}
L_{csl} = \sum_{j=1}^{n-1}\sum_{k=j+1}^n \text{sim}(c_{f_j},c_{f_k}) = \sum_{j=1}^n\sum_{k=1,k\neq j}^n \frac{c_{f_j}^T c_{f_k}}{\lVert c_{f_j} \rVert \lVert c_{f_k} \rVert}.
\end{equation}

\begin{equation} \label{eq:infonce}
L_{InfoNCE} = -\log\Bigg(\frac{\exp\big(\text{sim}(z_i, z_j)/\tau
    \big)}{\sum_{k=1,i\neq k}^N\exp\big(\text{sim}(z_i,z_k)/\tau\big)}\Bigg),
\end{equation}
where 
$\tau$ is the temperature hyperparameter.

\subsubsection{Fine-tuning Stage}

Fine-tuning initializes $f(\cdot)$ and $E(\cdot)$ with pretrained weights. A codebook generation module and an ECOC decoder are added.
The codebook generation module creates a codebook $C=\{\tilde{c}^{1},...,\tilde{c}^{K}\}$ using labeled data $D_{cg}$. The average codeword $\tilde{c}^{k}$ for class $k$ is:
\begin{equation}\label{eq:codeword_def}
    \tilde{c}^{k}=\frac{1}{N_k}\sum^N_{i=1}c^{i}\cdot1\{y_i = k\},
\end{equation}
where $N_k$ is the count of instances in class $k$.

Row separation loss ($L_{rsl}$, Eq.~\ref{eq:rsl_def}) ensures the distance between different class codewords.
\begin{equation}\label{eq:rsl_def}
L_{rsl} = -\log\Bigg(\frac{\exp\big(\text{sim}(c^i, c^{i+}/\tau
\big)}{\sum_{k=1,k\neq i}^N\exp\big(\text{sim}(c^i,\tilde{c}^k)/\tau\big)}\Bigg),
\end{equation}
where $c^{i+}$ represents codewords predicted in the same class as $x_i$.

The ECOC decoder computes class probability $p_k^i$ by:
\begin{equation}\label{eq:prob_def}
    p_k^i = \frac{\exp(\text{sim}(c^i,\tilde{c}^k))}{\sum_{j=1}^K \exp(\text{sim}(c^i,\tilde{c}^j))}
\end{equation}

\subsubsection{Loss Function Design}
\label{sec:loss_design}

In addition to $L_{csl}$ and $L_{rsl}$, ACL-PF uses two other losses, cross-entropy loss $L_{ce}$ (Eq.~\ref{eq:ce_def}) and Hinge loss $L_{hl}$ (Eq.~\ref{eq:hinge_def}), to measure prediction error and codeword distance.

\begin{equation}\label{eq:ce_def}
    L_{ce} = -\sum_{k=1}^K y_i\log(p^i_k)\end{equation}
\begin{equation}\label{eq:hinge_def}
    L_{hl} = \max\big(0.5 - \text{sim}(c^i, \tilde{c}^i), 0\big)
\end{equation}

The total losses for pretraining and fine-tuning:

\begin{equation}\label{eq:pretrain-loss}
L_{pretrain} = L_{infonce} + \lambda L_{csl}
\end{equation}
\begin{equation}\label{eq:finetune-loss}
L_{finetune} = L_{ce} + L_{hl} + L_{rsl}
\end{equation}

\subsection{ACL-CFPC: Co-Fine-tuning Model and Codebook with Pretrained Codebook} \label{cotrain_model&codebook}

\begin{figure*}[tb]
\centering
\includegraphics[width=\textwidth]{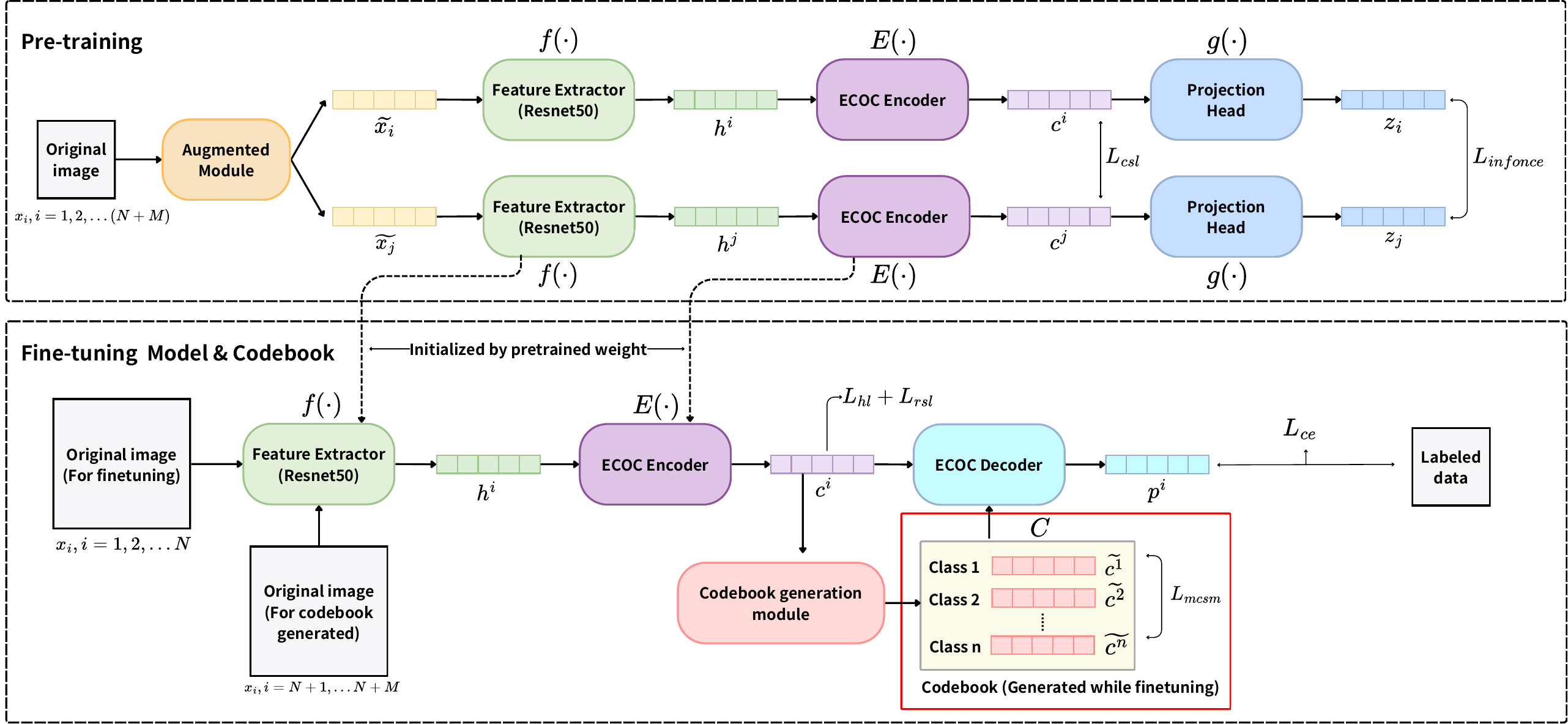}
\caption[ACL-CFPC Network Architecture]{ACL-CFPC network architecture. Fine-tuning adds MCSML and dynamically generates the codebook. Red boxes indicate differences from ACL-PF.}
\label{fig:model2_arch}
\end{figure*}

ACL-CFPC (Figure~\ref{fig:model2_arch}) modifies ACL-PF by adding Maximum Cosine Similarity Minimization Loss (Eq.~\ref{eq:mcsml-def}) for row separation and dynamically updating the codebook each batch using current weights.
\begin{equation}\label{eq:mcsml-def}
    L_{mcsm} = \max_{i \neq j} \text{sim}(\tilde{c}^i, \tilde{c}^j)
\end{equation}
Thus, the total fine-tuning loss (Eq.~\ref{eq:finetune-fixed-loss}) is

\begin{equation}\label{eq:finetune-fixed-loss}
    L_{finetune} = L_{ce} + L_{hl} + L_{rsl} + L_{mcsm}
\end{equation}

ACL-CFPC dynamically updates the codebook for each batch using current weights. This allows for better adaptation, potentially improving performance and robustness by keeping the codebook aligned with the model state.

\subsection{ACL-TFC: Training with Finetuned Codebook}\label{traing_with_finetune_codebook}

\begin{figure*}[tb]
\centering
\includegraphics[width=\textwidth]{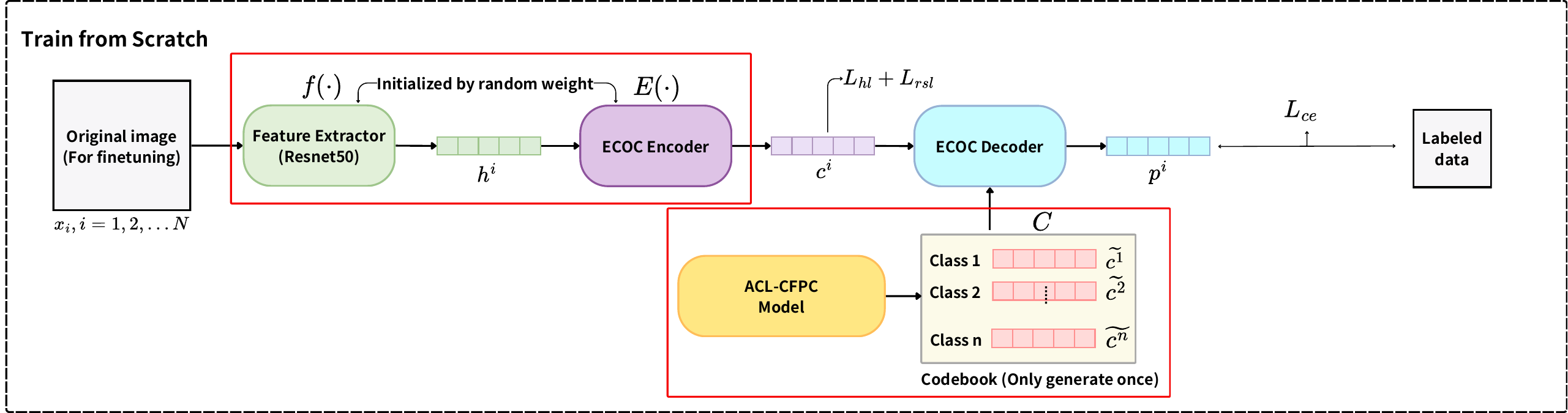}
\caption[ACL-TFC Network Architecture]{ACL-TFC network architecture. Trained with a fixed codebook from ACL-CFPC. Red boxes indicate differences.}
\label{fig:model3_arch}
\end{figure*}

ACL-TFC (Figure~\ref{fig:model3_arch}) uses ACL-CFPC as a pretraining step. It uses the refined codebook from ACL-CFPC as fixed targets and trains the model parameters from scratch. This uses the structured codebook for guided training, aiming to improve robustness and performance.

\subsection{Comparing the Proposed Models} \label{compared_with_three_method}

ACL-PF generates the codebook when pretraining is complete. ACL-CFPC dynamically adjusts it during fine-tuning. ACL-TFC uses a fixed codebook from ACL-CFPC and learns other parameters from scratch. ACL-PF benefits from initial guidance, ACL-CFPC from continuous refinement, and ACL-TFC from a fixed and structured codebook for robust training.

\section{Experiment}

\subsection{Datasets and Adversarial Examples}

Experiments use CIFAR-10, Fashion-MNIST, and GTSRB datasets.
Grayscale images are converted to 3-channel by replicating channels.
Each dataset is divided into training, testing, generation, and validation subsets (Table~\ref{tab:subdataset-amount}). The generation subset creates codebooks. The validation subset evaluates the best model for testing.

\begin{table}[tb]
    \caption[Number of Samples in Each Subset for the Four Datasets]{Sample counts in training, testing, generation, and validation subsets.}
    \label{tab:subdataset-amount}
    \centering
    \begin{tabular}{@{}c|cccc@{}}
    \toprule
    Dataset Name  & Train & Test & Generation & Validation    \\ \midrule
    CIFAR10       & 40000 & 8000 & 10000      & 2000          \\
    Fashion-MNIST & 48000 & 8000 & 12000      & 2000          \\
    GTSRB         & 21312 & 10104 & 5328       & 2526       \\ \bottomrule
    \end{tabular}
\end{table}

Robustness is tested using FGSM and PGD white-box attacks. Adversarial examples replace clean test data for evaluation. FGSM is a single-step attack maximizing loss. PGD is a stronger, iterative attack finding more effective perturbations.

\begin{table}[tbh]
    \caption{Performance of five models on open datasets under CLEAN, FGSM, and PGD conditions.}
    \label{tab:all-performance}
    \centering
    \begin{tabular}{c|c|c|c|c|c|c}
        \toprule
        & Dataset & Standard & SimCLR & ACL-PF & ACL-CFPC & ACL-TFC \\
        \midrule
        \multirow{3}{*}{CIFAR-10} & CLEAN & $86.27\%$ & $90.07\%$ & $90.1\%$ & $\bm{90.28\%}$ & $84.28\%$ \\
                                  & FGSM & $4.63\%$ & $19.96\%$ & $39.81\%$ & $\bm{40.77\%}$ & $21.94\%$ \\
                                  & PGD & $0\%$ & $0.06\%$ & $2.4\%$ & $3.21\%$ & $\bm{14.51}\%$ \\
        \midrule
        \multirow{3}{*}{Fashion-MNIST} & CLEAN & $\bm{92.7}\%$ & $92.41\%$ & $92.65\%$ & $92.64\%$ & $92.42\%$ \\
                                       & FGSM & $29.15\%$ & $39.54\%$ & $46.83\%$ & $53.45\%$ & $\bm{82.3}\%$ \\
                                       & PGD & $7.75\%$ & $3.77\%$ & $10.82\%$ & $12.65\%$ & $\bm{35.09}\%$ \\
        \midrule
        \multirow{3}{*}{GTSRB} & CLEAN & $\bm{92}\%$ & $90.67\%$ & $90.98\%$ & $89.67\%$ & $57.85\%$ \\
                               & FGSM & $36.81\%$ & $45.89\%$ & $52.26\%$ & $\bm{52.65}\%$ & $33.25\%$ \\
                               & PGD & $1.79\%$ & $7.76\%$ & $3.27\%$ & $4.67\%$ & $\bm{12.96}\%$ \\
        \bottomrule
    \end{tabular}
\end{table}

\subsection{Compared Baselines}\label{model_intro&eval}

The proposed models are compared with two baselines: Standard and SimCLR~\cite{chen2020simple}, which are a supervised learning baseline and a contrastive pretraining plus fine-tuning strategy where the fine-tuning architecture matches Standard.

\subsection{Accuracies on Clean and Attacked Datasets}\label{result_and_analysis}

The models are evaluated under three conditions: clean data, FGSM attack, and PGD attack. FGSM/PGD measure adversarial resilience.

ACL models show improved robustness, outperforming the baselines in FGSM and PGD attacks (Table~\ref{tab:all-performance}). Although Standard sometimes leads in clean data, ACL models are competitive. Among ACL models, ACL-CFPC and ACL-TFC generally show better attack robustness, although ACL-TFC's clean accuracy is lower. ACL-CFPC offers a better balance for clean and attacked data compared to ACL-TFC. Standard model's accuracy drops significantly under strong PGD attack. ACL-TFC best mitigates the PGD performance drop.

\section{Conclusion} \label{sec:disc}

We introduce models integrating SimCLR architecture and ECOC principles for automatic codebook learning, thereby removing the reliance on pre-designed codebooks. Experimental results demonstrate that the learned codebooks are robust against adversarial attacks such as FGSM and PGD. This facilitates the generation of dataset-specific codebooks, enhancing the flexibility and adaptability of ECOC design.

Although the proposed ACL models exhibit higher resilience to attacks compared to baseline models, there is still room for improvement. First, since ACL and adversarial training are not mutually exclusive, existing adversarial training methods can be combined with the codebooks obtained from the ACL models for joint training. This hybrid approach can potentially leverage the strengths of both methods further to enhance the model's robustness against adversarial attacks. Second, our study uses the SimCLR framework for contrastive learning to automatically learn codebooks. Future research could explore the integration of more advanced contrastive learning frameworks, such as MoCo, BYOL, or SwAV, which might provide more robust feature representation capabilities and improve model accuracy and robustness against attacks.

\section*{Acknowledgement and GenAI Usage Disclosure}

We acknowledge support from National Science and Technology Council of Taiwan under grant number 113-2221-E-008-100-MY3. We thank to National Center for High-performance Computing (NCHC) of National Applied Research Laboratories (NARLabs) in Taiwan for providing computational and storage resources.
The authors used ChatGPT and Gemini to improve language and readability. The authors reviewed and edited the content as needed and take full responsibility for the content of the publication.

\bibliographystyle{unsrt}  
\bibliography{ref}  

\end{document}